\documentclass[10pt,twocolumn,letterpaper]{article}

\usepackage[final]{cvpr}      
\usepackage{tablefootnote}
\usepackage{times}
\usepackage{epsfig}
\usepackage{graphicx}
\usepackage{amsmath}
\usepackage{amssymb}
\usepackage{multirow}
\usepackage{pifont}
\usepackage{booktabs}
\usepackage{multirow}

\usepackage[pagebackref=true,breaklinks=true,colorlinks,bookmarks=false]{hyperref}
\usepackage[table,xcdraw]{xcolor}
\usepackage{enumitem}
\usepackage{wrapfig}
\usepackage{caption,subcaption}

\definecolor{baselinecolor}{gray}{.9}
\newcommand{\authorskip}{\hspace{5mm}}

\def\cvprPaperID{1740} 
\def\confName{CVPR}
\def\confYear{2023}

\begin{document}

\title{Token Contrast for Weakly-Supervised Semantic Segmentation}

\author{Lixiang Ru$^{1}$ \authorskip Heliang Zheng$^{2}$ \authorskip Yibing Zhan$^{2}$ \authorskip Bo Du$^{1}$\thanks{Corresponding author. This work was done when Lixiang Ru was a research intern at JD Explore Academy.}
\\
$^{1}$ School of Computer Science, Wuhan University, China.\\
$^{2}$ JD Explore Academy, China
\\
{\tt\small {\{rulixiang, dubo\}@whu.edu.cn}\authorskip {\{zhengheliang,zhanyibing\}}@jd.com}
}
\maketitle

\begin{abstract}
  Weakly-Supervised Semantic Segmentation (WSSS) using image-level labels typically utilizes Class Activation Map (CAM) to generate the pseudo labels. Limited by the local structure perception of CNN, CAM usually cannot identify the integral object regions. Though the recent Vision Transformer (ViT) can remedy this flaw, we observe it also brings the over-smoothing issue, \ie, the final patch tokens incline to be uniform. In this work, we propose Token Contrast (ToCo) to address this issue and further explore the virtue of ViT for WSSS. Firstly, motivated by the observation that intermediate layers in ViT can still retain semantic diversity, we designed a Patch Token Contrast module (PTC). PTC supervises the final patch tokens with the pseudo token relations derived from intermediate layers, allowing them to align the semantic regions and thus yield more accurate CAM. Secondly, to further differentiate the low-confidence regions in CAM, we devised a Class Token Contrast module (CTC) inspired by the fact that class tokens in ViT can capture high-level semantics. CTC facilitates the representation consistency between uncertain local regions and global objects by contrasting their class tokens. Experiments on the PASCAL VOC and MS COCO datasets show the proposed ToCo can remarkably surpass other single-stage competitors and achieve comparable performance with state-of-the-art multi-stage methods. Code is available at \url{https://github.com/rulixiang/ToCo}.

\end{abstract}

\section{Introduction}
\par To reduce the expensive annotation costs of deep semantic segmentation models, weakly-supervised semantic segmentation (WSSS) is proposed to predict pixel-level predictions with only weak and cheap annotations, such as image-level labels \cite{ahn2018learning}, points \cite{bearman2016s}, scribbles \cite{zhang2021dynamic} and bounding boxes \cite{lee2021bbam}. Among all these annotation forms, the image-level label is the cheapest and contains the least information. This work also falls in the field of WSSS using only image-level labels.

\par Prevalent works of WSSS using image-level labels typically derive Class Activation Map (CAM) \cite{zhou2016learning} or its variants \cite{selvaraju2017grad} as pseudo labels. The pseudo labels are then processed with alternative refinement methods \cite{ahn2019weakly,ahn2018learning} and used to train regular semantic segmentation models. However, CAM is usually flawed since it typically only identifies the most discriminative semantic regions, severely weakening the final performance of semantic segmentation \cite{ahn2019weakly,wang2020self,jiang2022l2g}. The recent works \cite{gao2021ts,ru2022learning,xu2022multi} show one reason is that previous methods usually generate CAM with CNN, in which convolution only perceives local features and fails to activate the integral object regions. To ameliorate this problem and generate more accurate pseudo labels for WSSS, these works propose solutions based on the recent Vision Transformer (ViT) architecture \cite{dosovitskiy2020image}, which inherently models the global feature interactions with self-attention blocks.

\begin{figure}[tbp]
  \centering
  \includegraphics[width=0.45\textwidth,]{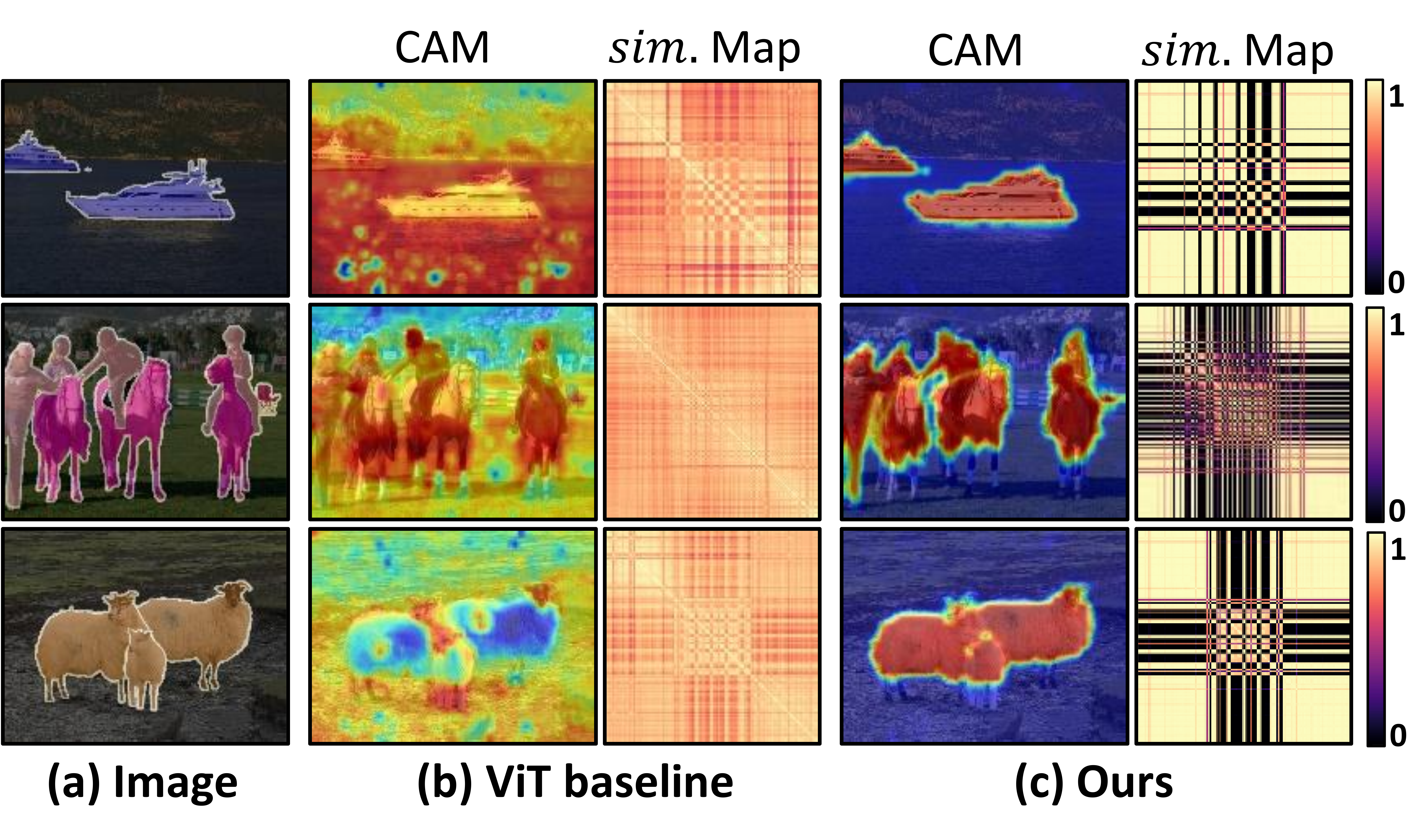}
  \vspace{-0.2cm}
  \caption{\textbf{The generated CAM and the pairwise cosine similarity of patch tokens ($sim.$ map).} Our method can address the over-smoothing issue well and produce accurate CAM. Here we use ViT-Base.}
  \label{fig_teaser}
  \vspace{-0.6cm}
\end{figure}

\par However, as demonstrated in \cite{park2021vision,wang2021anti}, self-attention in ViT is essentially a low-pass filter, which inclines to reduce the variance of input signals. Therefore, stacking self-attention blocks is equivalent to repeatedly performing spatial smoothing operations, which encourages the patch tokens in ViT to be uniform \cite{shi2021revisiting,gong2021vision}, \ie, over-smoothing. We observe that the over-smoothing issue particularly impairs the WSSS task, since CAM used to derive pseudo labels relies on the output features (\ie patch tokens). As shown in Figure~\ref{fig_teaser}, due to over-smoothing, the pairwise cosine similarities of the patch tokens are close to $1$, suggesting the learned representations of different patch tokens are almost uniform. The generated CAM thus tends to assign different image regions with the monotonous semantic label. Though several recent works have explored the ViT architecture for WSSS \cite{xu2022multi,sun2021getam,ru2022learning}, they typically overlook the over-smoothing issue of patch tokens, leaving this problem unresolved.

\par In this work, we empirically observe that ViT smooths the patch tokens progressively, \ie the learned representations in intermediate layers can still preserve the semantic diversity. Therefore, we propose a Patch Token Contrast (PTC) module to address the over-smoothing issue by supervising the final patch tokens with intermediate layer knowledge. Specifically, in the PTC module, we simply add an additional classifier in an intermediate layer to extract the auxiliary CAM and the corresponding pseudo pairwise token relations. By supervising the pairwise cosine similarities of final patch tokens with the pseudo relations, PTC can finely counter the over-smoothing issue and thus produce high-fidelity CAM. As shown in Figure~\ref{fig_teaser}, our method can generate CAM that aligns well with the semantic object regions. The pairwise cosine similarities also coincide with the corresponding semantics. In addition, to further differentiate the uncertain regions in generated CAM, inspired by the property that the class token in ViT can inherently aggregate high-level semantics \cite{caron2021emerging,gao2021ts}, we also propose a Class Token Contrast (CTC) module. In CTC, we first randomly crop local images from uncertain regions (background regions), and minimize (maximize) the representation difference between the class tokens of local and global images. As a result, CTC can facilitate the local-to-global representation consistency of semantic objects and the discrepancy between foreground and background, benefiting the integral and accurate object activation in CAM. Finally, based on the proposed PTC and CTC, we build Token Contrast (ToCo) for WSSS and extend it to the single-stage WSSS framework \cite{ru2022learning}.

\par Overall, our contributions in this work include the following aspects.
\begin{itemize}[noitemsep,nolistsep,leftmargin=*]
  \item We propose Patch Token Contrast (PTC) to address the over-smoothing issue in ViT. By supervising the final tokens with intermediate knowledge, PTC can counter the patch uniformity and significantly promote the quality of pseudo labels for WSSS.
  \item We propose Class Token Contrast (CTC), which contrasts the representation of global foregrounds and local uncertain regions (background) and facilitates the object activation completeness in CAM.
  \item The experiments on the PASCAL VOC \cite{everingham2010pascal} and MS COCO dataset \cite{lin2014microsoft} show that the proposed ToCo can significantly outperform SOTA single-stage WSSS methods and achieve comparable performance with multi-stage competitors.
\end{itemize}
\vspace{-2mm}
\section{Related Work}
\noindent\textbf{Weakly-Supervised Semantic Segmentation.} Weakly-Supervised Semantic Segmentation (WSSS) using image-level labels typically generates CAM as the initial pseudo labels. A typical drawback of CAM is that it usually only activates the most discriminative object regions. To address this drawback, recent works proposed various training schemes, such as erasing \cite{wei2017object}, online attention accumulation \cite{jiang2019integral} and cross-image semantic mining \cite{sun2020mining}. \cite{ru2022weakly,chang2020weakly,wang2020self} propose to leverage auxiliary tasks to regularize the training objective, such as visual words learning \cite{ru2022weakly}, sub-category exploration \cite{chang2020weakly}, and scale in-variance regularization \cite{wang2020self}. \cite{lee2021railroad,yao2021non} utilize extra saliency maps as supervision to suppress the background regions and mine the non-salient objects. \cite{xie2022cross,su2021context,lee2022weakly} counter the problem of semantic co-occurrence via distilling knowledge from CLIP \cite{radford2021learning}, decoupling object context \cite{su2021context} and comparing the out-of-distribution images \cite{lee2022weakly}, respectively. \cite{zhou2022regional,du2022weakly,chen2022self} contrast the pixel and prototype representations to encourage the integral activation of the object regions. Typically, these methods are built upon the CNN network, inheriting the aforementioned locality flaw. In this work, we explore ViT for WSSS to avoid this drawback and achieve integral object activation.

\noindent\textbf{Vision Transformer for WSSS.} Vision Transformer (ViT) has achieved great success in various vision tasks \cite{dosovitskiy2020image,xie2021segformer,carion2020end,xu2023demt,lan2022learning}. Some recent works also introduce ViT to WSSS \cite{gao2021ts,sun2021getam,ru2022learning,xu2022multi}. Inspired by the property that the class token in ViT can capture the foreground information \cite{caron2021emerging}, TS-CAM \cite{gao2021ts} extracts the class-agnostic attention map and couples it with the naive semantic-aware CAM. \cite{sun2021getam} proposes to derive the gradients of the attention map and extracts the attention of {class token} \wrt other tokens as the class-specific maps. MCTformer \cite{xu2022multi} embeds multiple {class tokens} and enforces them learning the activation maps of different classes. AFA \cite{ru2022learning} proposes to learn reliable semantic affinity from the attention blocks to refine the initial coarse labels. However, these methods typically overlook the over-smoothing issue of ViT. Besides, they need to modify the ViT architecture or extract the costly gradients at the inference stage. In this work, we propose ToCo to solve the over-smoothing issue without modification to architecture and further unlock the potential of ViT on the WSSS task.

\section{Preliminaries}
In this section, we briefly introduce the preliminary knowledge of class activation map (CAM), Vision Transformer (ViT), and over-smoothing in ViT.
\subsection{Class Activation Map}
\par CAM \cite{zhou2016learning} is proposed to identify the activated regions when a classification network predicts an image. Due to its efficacy and simplicity, CAM has been widely used to generate the initial pseudo labels for WSSS. Specifically, given an image, its feature maps $\mathbf{F}\in \mathbb{R}^{hw\times d}$ are extracted with a classification network (CNN or ViT), where $hw$ and $d$ denote the number of spatial and channel dimension, respectively. The CAM is computed by weighting and summing the feature maps with the weights $\mathbf{W} \in\mathbb{R}^{c\times d}$ in the classification layer, where $c$ is the number of semantic classes. The \texttt{relu} function and \texttt{max} normalization are then applied to eliminate the negative activation and scale the CAM to $[0,1]$. Therefore, CAM for class $c$ is thus calculated as
\begin{equation}\small
  \mathtt{CAM}_c(\mathbf{F},\mathbf{W}) = \frac{\mathtt{relu}(\mathbf{M}_c)}{\max(\mathtt{relu}(\mathbf{M}_c))}, \mathbf{M}_c=\sum_{i}\mathbf{W}_{c,i}\mathbf{F}_{:,i}.
  \label{eq_cam}
\end{equation}
A background threshold $\beta$ is usually used to differentiate the background and foreground regions.

\subsection{Vision Transformer \& Over-smoothing}
\label{sec_trans_smoothing}
\begin{figure}[tbp]
  \centering
  \includegraphics[width=0.35\textwidth,]{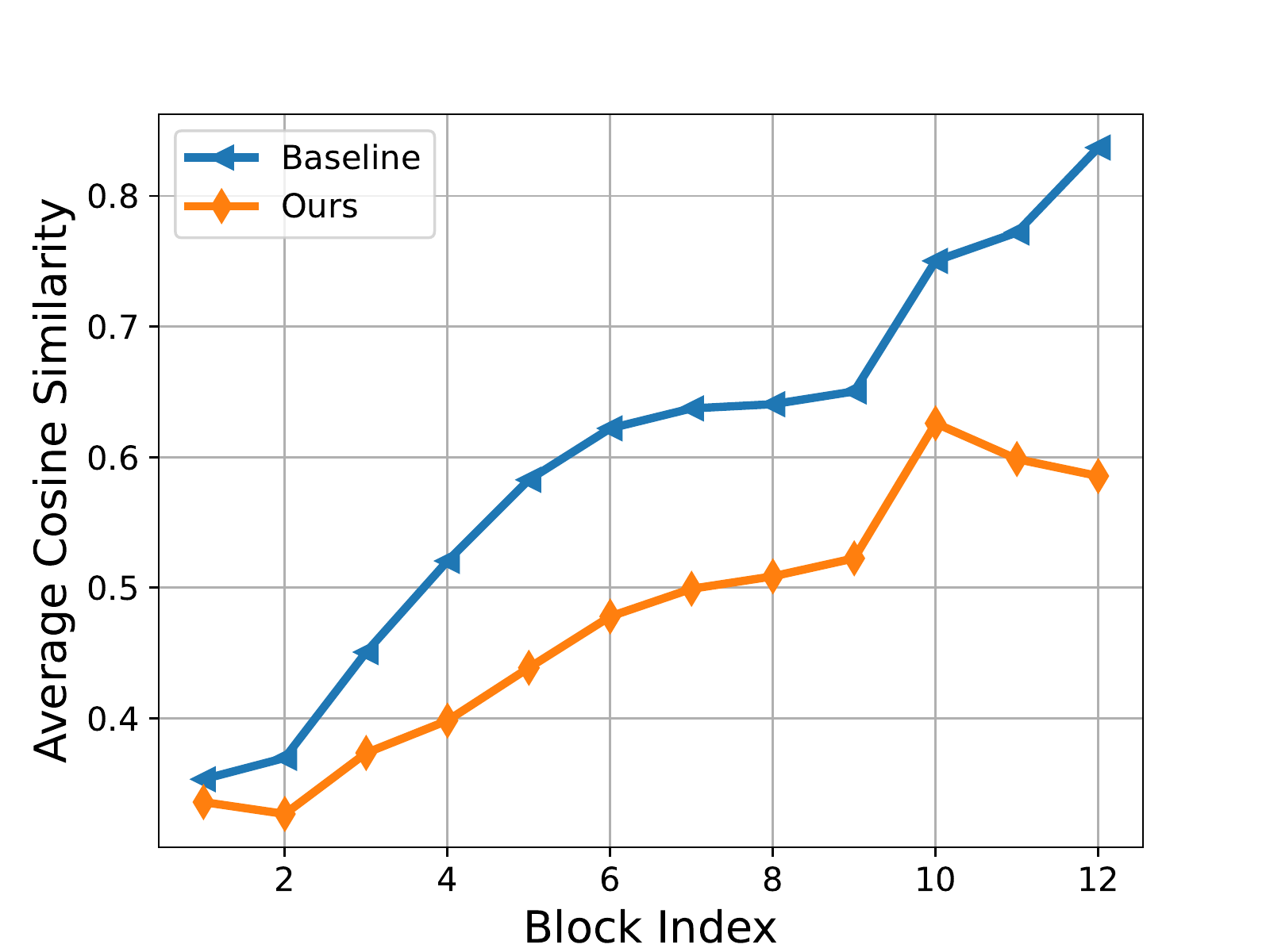}
  \caption{\textbf{The average pairwise cosine similarity of patch tokens in each Transformer block.} The cosine similarity is computed on the VOC \texttt{train} set. Here we use the ViT-Base (ViT-B) \cite{dosovitskiy2020image} architecture which includes 12 Transformer blocks.}
  \label{fig_voc_cos_sim}
  \vspace{-0.4cm}
\end{figure}

\par An ViT firstly splits an image into patches to form the initial patch tokens. Then the patch tokens are concatenated with an extra learnable class token and feed into the Transformer encoder to obtain the final patch and class tokens. As the key component, in each Transformer block, the multi-head self-attention (MHSA) is used to perform the global feature interaction. However, due to the low-pass property of self-attention \cite{park2021vision,wang2021anti}, after multiple Transformer blocks, the output patch tokens incline to be uniform, which severely affects the CAM according to Equation~\ref{eq_cam}.
\par In Figure~\ref{fig_voc_cos_sim}, we visualized the pairwise cosine similarity of the patch tokens generated in each Transformer block. Figure~\ref{fig_voc_cos_sim} shows that the patch tokens in the late layers are highly similar, while the early layers can still preserve the semantic diversity. This observation motivates us to address the over-smoothing issue by supervising the final layer tokens with knowledge from intermediate layers.

\section{Methodology}
\par This section elaborates on the proposed method, \ie Token Contrast (ToCo) for WSSS. We first introduce the overall framework of ToCo. Then the Patch Token Contrast (PTC) and Class Token Contrast (CTC) are proposed to address the over-smoothing issue and further exploit the virtue of ViT for WSSS, respectively. Finally, we present the training objective of ToCo and how to plug it into the single-stage WSSS framework.
\begin{figure*}[!tbp]
  \centering
  \includegraphics[width=0.8\textwidth]{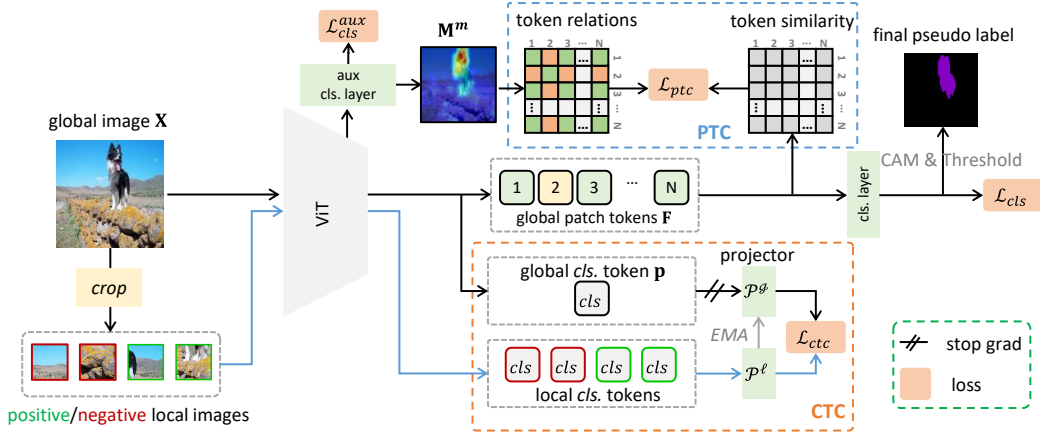}
  \caption{\textbf{The overall framework of ToCo}. ToCo firstly uses an additional classification layer ($cls.$ layer) to produce the auxiliary CAM ($aux.$ CAM). In the PTC module, the $aux.$ CAM is used to derive token relations and supervise the pairwise token similarities of final patch tokens to address the over-smoothing issue. In the CTC module, the class tokens of the negative/positive images will be projected and contrasted with the global class token to further differentiate the low-confidence regions in CAM. The pseudo label is generated with the final CAM.}
  \label{fig_toco}
  \vspace{-4mm}
\end{figure*}

\subsection{Overview}
\label{sec_overall}
\par As illustrated in Figure~\ref{fig_toco}, ToCo uses an auxiliary classification layer in the ViT encoder to produce the auxiliary CAM. The auxiliary CAM is subsequently leveraged to generate the auxiliary pseudo labels and guide the PTC module. Meanwhile, it's also used to produce proposals to crop positive and negative local images for the CTC module. The final CAM is obtained with a classification layer and used to generate the final pseudo labels.

\subsection{Patch Token Contrast}
\label{sec_ptc}
\par The objective of the Patch Token Contrast (PTC) module is to address the over-smoothing issue of the final patch tokens. As aforementioned, since the intermediate layers can still preserve the semantic diversity of patch tokens, in PTC, we leverage knowledge from the intermediate layer, \ie reliable pairwise token relations in Figure~\ref{fig_toco}, to supervise the final patch tokens.

\par Specifically, an input image $\mathbf{X}$ is firstly tokenized to construct initial patch tokens and then passed through the Transformer encoder. For a given intermediate layer, the output patch tokens are denoted as $\mathbf{F}^{m}\in\mathbb{R}^{n\times d}$, where $n$ and $d$ denotes the number of tokens and the feature dimension, respectively. As shown in Figure~\ref{fig_toco}, to extract the semantic-aware knowledge, we add an auxiliary classification layer to perform classification and generate CAM. In practice, we observe that not all intermediate layers produce satisfactory CAM for supervising PTC, since late layers incline to smooth the patch tokens while early layers may fail to capture high-level semantics. The choice of the intermediate layer will be discussed in Section~\ref{sec_ablation}.

\par In the auxiliary classification head, the patch tokens $\mathbf{F}^{m}$ are firstly aggregated via global max-pooling (GMP) as suggested in \cite{ru2022learning} and then projected with a fully-connected layer parameterized with $\theta^{m}$ to calculate the auxiliary classification loss $\mathcal{L}^{m}_{cls}$. Therefore, the auxiliary CAM is computed as
\begin{equation}\small
  \mathbf{M}^{m} =\mathtt{CAM}(\mathbf{F}^{m}, \theta^{m}).
  \label{eq_cam_aux}
\end{equation}
We then use two background thresholds $\beta_l, \beta_h$ ($0<\beta_l<\beta_h<1$) to segment $\mathbf{M}^{m}$ to the pseudo token label $\mathbf{Y}^{m}$, which consists of reliable foreground, background and uncertain regions.

\noindent \textbf{Patch Token Contrast Loss.} The generated token label $\mathbf{Y}^{m}$ is used to derive reliable pairwise relations for supervising the final patch tokens. Specifically, if two tokens share the same semantic label, they are labeled as positive pairs; otherwise, they are labeled as negative pairs. In addition, to ensure reliability, we only consider two tokens that both belong to the reliable foreground or background regions and ignore the uncertain regions. To remedy the over-smoothing issue, we maximize the similarity of two final patch tokens that belong to positive pairs and minimize the similarity otherwise. Let $\mathbf{F} \in \mathbb{R}^{n\times d}$ be the final layer patch tokens, the loss function for the PTC module is then constructed as
\begin{equation} \small
  \begin{aligned}
    \mathcal{L}_{ptc} & = \frac{1}{{N}^+}\sum_{\mathbf{Y}_i = \mathbf{Y}_j}(1-\mathtt{CosSim}(\mathbf{F}_i,\mathbf{F}_j)) \\
                      & +\frac{1}{{N}^-}\sum_{\mathbf{Y}_i \neq \mathbf{Y}_j}\mathtt{CosSim}(\mathbf{F}_i,\mathbf{F}_j),
  \end{aligned}
  \label{eq_loss_ptc}
\end{equation}
where $\mathtt{CosSim}(\cdot,\cdot)$ computes the cosine similarity and ${N}^+$/${N}^-$ counts the number of positive/negative pairs. However, minimizing the original cosine similarity cannot ensure the diversity \cite{gong2021vision,chen2022principle}, since a token pair with minus cosine similarity (\eg, $-1$) could be highly correlated. Therefore, in practice, in Equation~\ref{eq_loss_ptc}, we use the absolute cosine similarity instead of the original form. By minimizing Equation~\ref{eq_loss_ptc}, the representations of positive tokens are encouraged to be more consistent, while the negative tokens pairs are more discriminative, so that the over-smoothing issue can be well addressed. 

\subsection{Class Token Contrast}
\label{sec_ctc}
Addressing the over-smoothing with PTC can drive ViT to generate compelling CAM and pseudo labels. However, inevitably, there are still some less discriminative object regions that are hard to differentiate in CAM. Inspired by the property that class tokens in ViT can aggregate the high-level semantics \cite{gao2021ts,caron2021emerging}, we design a Class Token Contrast (CTC) module to facilitate the representation consistency between the local non-salient regions and the global object, which can further enforce more object regions to be activated in CAM.

\par
\begin{figure}[tbp]
  \centering
  \includegraphics[width=0.45\textwidth,]{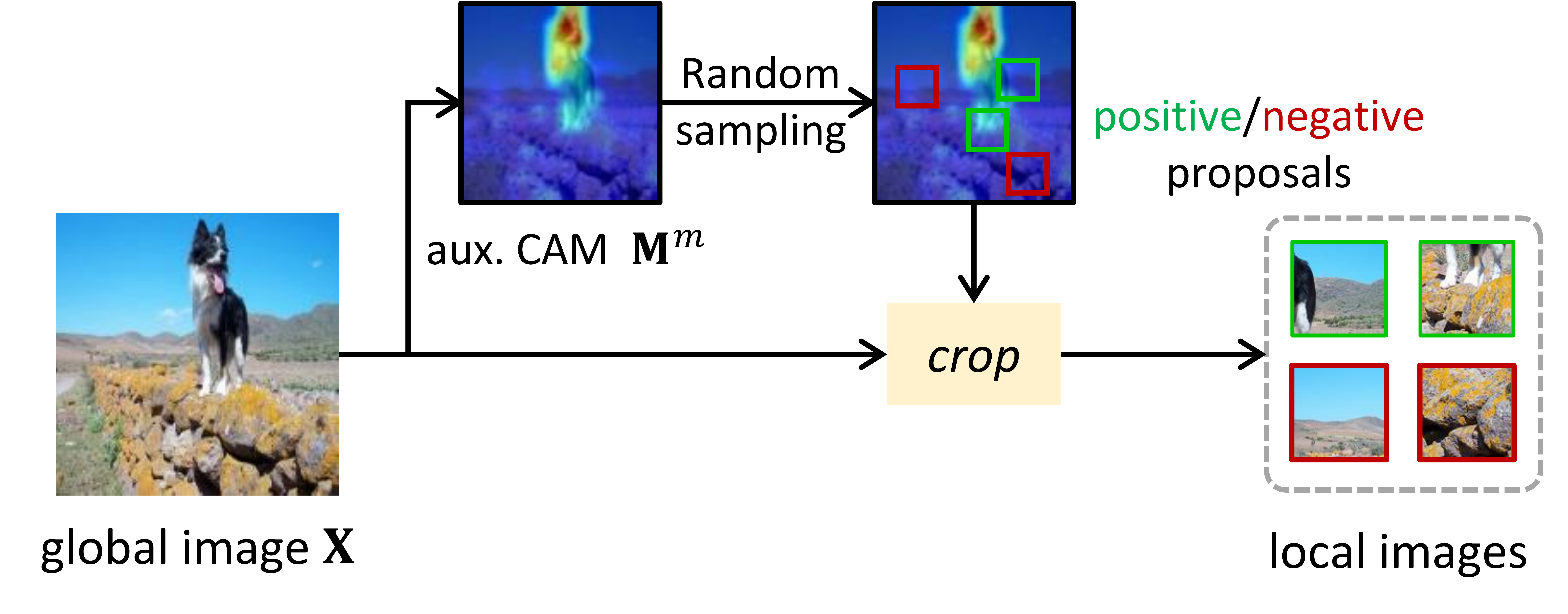}
  \caption{\textbf{Illustration of the crop method in Figure~\ref{fig_toco}.}}
  \label{fig_toco_crop}
  \vspace{-0.4cm}
\end{figure}

 As illustrated in Figure~\ref{fig_toco_crop}, given an image, we first randomly crop the local images from the uncertain regions specified by its auxiliary CAM. Since the class token in ViT captures the information of semantic objects \cite{caron2021emerging,gao2021ts}, the class tokens of global and local images aggregate the information of the global and the local objects, respectively. By minimizing the difference between global and local class tokens, the representation of entire object regions can be more consistent.

To counter the case that the cropped local images may contain few/no foreground objects, as shown in Figure~\ref{fig_toco}, we also crop some local images from background regions. By maximizing the difference between class tokens of global image and local background regions, the foreground-background discrepancy can be also facilitated. In practice, we randomly crop a fixed number of local images and assign them as positive (from uncertain regions) or negative (from background regions) with the guidance of $\mathbf{Y}^{m}$ in Section~\ref{sec_ptc}.

\par Specifically, the global and local class tokens are first passed through the projection head $\mathcal{P}^g$ and $\mathcal{P}^l$, respectively, which consist of linear layers and an L2 normalization layer. Assuming $\mathbf{p}$ denotes the projected global class token, and $\mathcal{Q}^+$/$\mathcal{Q}^-$ denotes the set of projected local class tokens cropped from uncertain/background regions, the objective of CTC is to minimize/maximize the difference between $\mathbf{p}$ and the local class tokens in $\mathcal{Q}^+$/$\mathcal{Q}^-$. Here we use the InfoNCE loss \cite{oord2018representation} as the objective, \ie
\begin{equation}\small
  \mathcal{L}_{ctc} = \frac{1}{N^+}\sum_{\mathbf{q^+}}\log{\frac{e^{(\mathbf{p}^\top\mathbf{q^+}/\tau)}}{e^{(\mathbf{p}^\top\mathbf{q^+}/\tau)}+\sum_{\mathbf{q^-}}{e^{(\mathbf{p}^\top\mathbf{q^-}/\tau)}+\epsilon}}},
  \label{eq_loss_ctc}
\end{equation}
where $\mathbf{q}^+\in\mathcal{Q}^+$, $\mathbf{q}^-\in\mathcal{Q}^-$, $N^+$ counts the number of $\mathcal{Q}^+$, $\tau$ is the temperature factor, and $\epsilon$ is a small positive value for stability. It's noted that CTC aims to enforce the local view representation to align the global view's. Therefore, we stop the gradient of the projection head $\mathcal{P}^g$. To update $\mathcal{P}^g$, we use the exponential moving average (EMA), \ie, $\theta^g \leftarrow \rho\theta^g+(1-\rho)\theta^l$, where $\rho$ is the momentum factor, $\theta^g$ and $\theta^l$ are parameters from  $\mathcal{P}^g$ and $\mathcal{P}^l$, respectively.

\subsection{ToCo for WSSS}
\label{sec_wsss}

\begin{figure*}[t]
  \centering
  \includegraphics[width=0.9\linewidth]{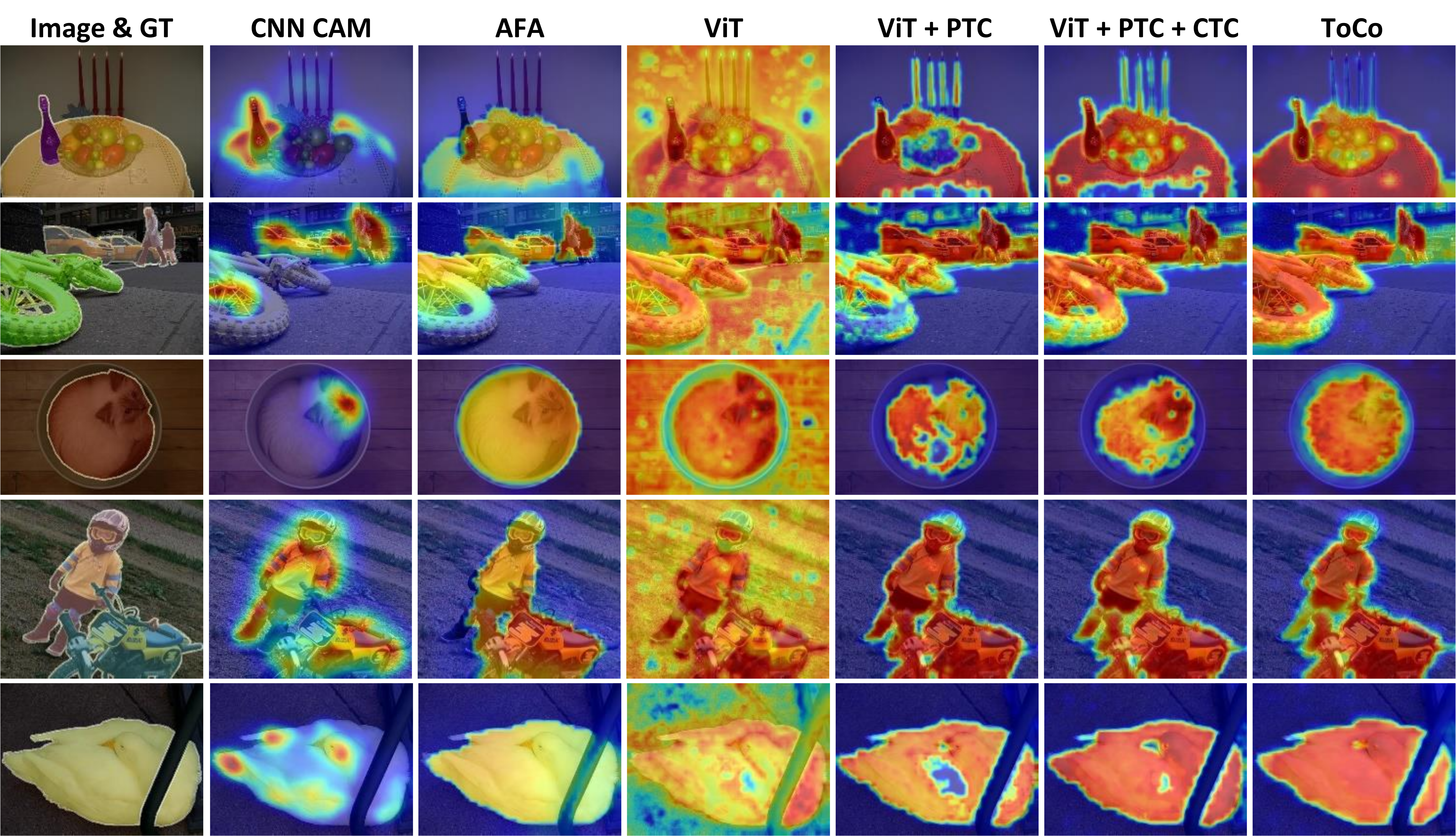}
  \caption{\textbf{Visualization of CAM.} From left to right, the CAM is generated with CNN baseline, AFA \cite{ru2022learning}, ViT baseline, ViT with PTC, ViT with PTC and CTC, and the proposed ToCo. }
  \label{fig_toco_cam}
  \vspace{-4mm}
\end{figure*}

\noindent\textbf{Training Objective.} As shown in Figure~\ref{fig_toco}, except the token contrast losses $\mathcal{L}_{ptc} $ and $\mathcal{L}_{ctc}$, the training objective of the proposed ToCo also includes the classification loss $\mathcal{L}_{cls}$ and auxiliary classification loss $\mathcal{L}_{cls}^{m}$. Following the common practice, we use the multi-label soft margin loss for both $\mathcal{L}_{cls}$ and $\mathcal{L}_{cls}^{m}$. The optimization objective of ToCo is the weighted sum of these loss terms:

\begin{equation}\small
  \mathcal{L}_{toco} = \mathcal{L}_{cls}+\mathcal{L}^{m}_{cls}+\lambda_1\mathcal{L}_{ptc}+\lambda_2\mathcal{L}_{ctc}.
  \label{eq_toco_loss}
\end{equation}

\noindent\textbf{Single-Stage WSSS.} We plug the proposed ToCo into the single-stage WSSS framework. Specifically, the pseudo labels produced by ToCo are then refined with a pixel-adaptive refinement module (PAR) \cite{ru2022learning} to align the low-level semantic boundaries. The refined pseudo labels will be used to supervise the segmentation decoder. We use the common cross-entropy loss as the segmentation loss $\mathcal{L}_{seg}$. The overall training objective should thus include $\mathcal{L}_{seg}$, \ie $\mathcal{L} = \mathcal{L}_{toco}+\lambda_3\mathcal{L}_{seg}$. Following previous single-stage WSSS works \cite{ru2022weakly,pan2022learning}, we also use an extra regularization loss term \cite{tang2018regularized} to enforce the spatial consistency of the predicted segmentation masks.

\begin{table}[tbp]
  \centering


  \center
  \small
  {
    \begin{tabular}{l|c|ccc}
      \toprule
      {Method}                                           & Backbone        & \texttt{train} & \texttt{val}  \\\midrule
      RRM \cite{zhang2020reliability} \tiny AAAI'2020    & WR38            & --             & 65.4          \\
      1Stage \cite{araslanov2020single}  \tiny CVPR'2020 & WR38            & 66.9           & 65.3          \\
      AA\&LR \cite{zhang2021adaptive} \tiny ACM MM'2021  & WR38            & 68.2           & 65.8          \\
      SLRNet \cite{pan2022learning} \tiny IJCV'2022      & WR38            & 67.1           & 66.2          \\
      AFA \cite{ru2022learning} \tiny CVPR'2022          & MiT-B1          & 68.7           & 66.5          \\
      ViT-PCM \cite{ros2022max} \tiny ECCV'2022          & ViT-B$^\dagger$ & 67.7           & 66.0          \\
      ViT-PCM + CRF \cite{ros2022max} \tiny ECCV'2022    & ViT-B$^\dagger$ & 71.4           & 69.3          \\
      \rowcolor[HTML]{eaeaea}
      \textbf{ToCo}                                      & ViT-B           & 72.2           & 70.5          \\
      \rowcolor[HTML]{eaeaea}
      \textbf{ToCo$^\dagger$}                            & ViT-B$^\dagger$ & \textbf{73.6}  & \textbf{72.3} \\ \bottomrule
    \end{tabular}
    \caption{\textbf{Evaluation of pseudo labels}. $\dagger$ denotes using ImageNet-21k \cite{ridnik2021imagenet} pretrained parameters.}
    \label{tab_pseudo_label_s}%
  }
  \vspace{-6mm}
\end{table}

\section{Experiments}
\subsection{Experimental Settings}
\noindent\textbf{Datasets.} We evaluate the proposed method on the PASCAL VOC 2012 \cite{everingham2010pascal} and MS COCO 2014 dataset \cite{lin2014microsoft}. Following common practice, {VOC 2012} dataset is further augmented with the SBD dataset \cite{hariharan2011semantic}. The \texttt{train}, \texttt{val}, and \texttt{test} set of the augmented dataset consist of 10582, 1449, and 1456 images, respectively. For {COCO 2014} dataset, the \texttt{train} and \texttt{val} set consist of about 82k and 40k images, respectively. In the training stage, we only use image-level labels. By default, we report mIoU as the evaluation metric.

\noindent\textbf{Network Architectures.} We use the ViT-base (ViT-B) \cite{dosovitskiy2020image} as the backbone, which is initialized with ImageNet pretrained weights \cite{ridnik2021imagenet}. To ensure the backbone accepts input images of arbitrary size, the \texttt{pos\_embedding} will be resized to input size via bilinear interpolation. The projection heads in the CTC module, \ie $\mathcal{P}^g$ and $\mathcal{P}^l$ in Figure~\ref{fig_toco}, consist of 3 linear layers and an L2-normalization layer. The parameters in projection heads are randomly initialized. We use a simple segmentation head as the decoder, which consists of two $3\times 3$ convolutional layers (with a dilation rate of 5) and a $1\times 1$ prediction layer.

\noindent\textbf{Implementation Details.} We train ToCo with an AdamW optimizer. The learning rate linearly increases to $6e^{-5}$ in the first 1500 iterations and decays with a polynomial scheduler for later iterations. The warm-up and decay rates are set as $1e^{-6}$ and 0.9, respectively. For experiments on the VOC dataset, the batch size and total iterations are set as 4 and 20k, respectively. The crop size of global and local view images are $448^2$ and $96^2$, respectively. Besides, we follow the multi-crop and data augmentation strategy in \cite{caron2021emerging} for global and local views. By default, the background thresholds $(\beta_l, \beta_h)$ are set as (0.25, 0.7). The temperature factors $\tau$ in Equation~\ref{eq_loss_ctc} is 0.5. The momentum factor for the EMA process in the CTC module is set as 0.9. The weight factors $(\lambda_1, \lambda_2, \lambda_3)$ of the loss terms in Section~\ref{sec_wsss} are set as (0.2, 0.5, 0.1). In the inference stage, following the common practice in semantic segmentation \cite{chen2017deeplab}, we use multi-scale testing and dense CRF processing.
\par For the experiments on the COCO dataset, the network is trained for 80k iterations with a batch size of 8. The background thresholds $(\beta_l, \beta_h)$ are set as (0.25, 0.65), while other settings remain the same. The impact of hyper-parameters will be presented in Section~\ref{sec_ablation} and Supplementary Material.

\subsection{Experimental Results}

\begin{table}[tbp]
  \small
  \centering
  \setlength{\tabcolsep}{0.8mm}
  \begin{tabular}{l|c|c|cc|c}
    \toprule
                                                      & \multirow{2}{*}{$Sup.$}   & \multirow{2}{*}{$Net.$  } & \multicolumn{2}{c|}{\textbf{VOC}} & {\textbf{COCO}}                                                                                           \\ \cmidrule{4-6}
                                                      &                           &                           & \texttt{val}                      & \texttt{test}                                                                             & \texttt{val}  \\ \midrule
    \multicolumn{4}{l}{\cellcolor[HTML]{ffffff}\textbf{\textit{Multi-stage WSSS methods}}.}                                                                                                                                                                   \\
    RIB \cite{lee2021reducing} \tiny NeurIPS'2021     & $\mathcal{I}+\mathcal{S}$ & DL-V2                     & 70.2                              & 70.0                                                                                      & --            \\
    EPS \cite{lee2021railroad}  \tiny CVPR'2021       & $\mathcal{I}+\mathcal{S}$ & DL-V2                     & 71.0                              & 71.8                                                                                      & --            \\
    L2G \cite{jiang2022l2g} \tiny CVPR'2022           & $\mathcal{I}+\mathcal{S}$ & DL-V2                     & 72.1                              & 71.7                                                                                      & 44.2          \\
    RCA   \cite{zhou2022regional}   \tiny CVPR'2022   & $\mathcal{I}+\mathcal{S}$ & DL-V2                     & 72.2                              & 72.8                                                                                      & 36.8          \\
    Du \etal   \cite{du2022weakly}   \tiny CVPR'2022  & $\mathcal{I}+\mathcal{S}$ & DL-V2                     & 72.6                              & 73.6                                                                                      & --            \\
    RIB \cite{lee2021reducing} \tiny NeurIPS'2021     & $\mathcal{I}$             & DL-V2                     & 68.3                              & 68.6                                                                                      & 43.8          \\
    ReCAM \cite{chen2022class} \tiny CVPR'2022        & $\mathcal{I}$             & DL-V2                     & 68.4                              & 68.2                                                                                      & 45.0          \\
    VWL \cite{ru2022weakly}  \tiny IJCV'2022          & $\mathcal{I}$             & DL-V2                     & 69.2                              & 69.2                                                                                      & 36.2          \\
    W-OoD \cite{lee2022weakly} \tiny CVPR'2022        & $\mathcal{I}$             & WR38                      & 70.7                              & 70.1                                                                                      & --            \\
    MCTformer \cite{xu2022multi} \tiny CVPR'2022      & $\mathcal{I}$             & WR38                      & 71.9                              & 71.6                                                                                      & 42.0          \\
    ESOL \cite{li2022expansion} \tiny NeurIPS'2022    & $\mathcal{I}$             & DL-V2                     & 69.9                              & 69.3                                                                                      & 42.6          \\\midrule
    \multicolumn{4}{l}{\cellcolor[HTML]{ffffff}\textbf{\textit{Single-stage WSSS methods}}.}                                                                                                                                                                  \\
    RRM \cite{zhang2020reliability} \tiny AAAI'2020   & $\mathcal{I}$             & WR38                      & 62.6                              & 62.9                                                                                      & --            \\
    1Stage \cite{araslanov2020single} \tiny CVPR'2020 & $\mathcal{I}$             & WR38                      & 62.7                              & 64.3                                                                                      & --            \\
    AFA \cite{ru2022weakly} \tiny CVPR'2022           & $\mathcal{I}$             & MiT-B1                    & 66.0                              & 66.3                                                                                      & 38.9          \\
    SLRNet \cite{pan2022learning} \tiny IJCV'2022     & $\mathcal{I}$             & WR38                      & 67.2                              & 67.6                                                                                      & 35.0          \\
    \rowcolor[HTML]{eaeaea}
    \textbf{ToCo}                                     & $\mathcal{I}$             & ViT-B                     & 69.8                              & 70.5\tablefootnote{\url{http://host.robots.ox.ac.uk:8080/anonymous/KV9AQH.html}}          & 41.3          \\
    \rowcolor[HTML]{eaeaea}
    \textbf{ToCo$^\dagger$}                           & $\mathcal{I}$             & ViT-B$^\dagger$           & \textbf{71.1}                     & \textbf{72.2}\tablefootnote{\url{http://host.robots.ox.ac.uk:8080/anonymous/GNTBBZ.html}} & \textbf{42.3} \\ \bottomrule
  \end{tabular}
  \vspace{-1mm}
  \caption{\textbf{Semantic Segmentation Results}. $Sup.$ denotes the supervision type. $\mathcal{I}$: Image-level labels; $\mathcal{S}$: Saliency maps. $Net.$ denotes the backbone network (for single-stage methods) and the semantic segmentation network (for multi-stage methods). $\dagger$ denotes using ImageNet-21k \cite{ridnik2021imagenet} pretrained parameters.}
  \label{tab_sem_seg}
  \vspace{-6mm}
\end{table}

\noindent\textbf{{Pseudo Labels}.} We first visualize the generated CAM with ToCo in Figure~\ref{fig_toco_cam}. As shown in Figure~\ref{fig_toco_cam}, our method can remarkably produce more integral and accurate CAM than the CNN methods and the recent single-stage WSSS method, \ie AFA \cite{ru2022learning}. Compared to the ViT baseline, our ToCo also remedies the over-smoothing issue well.
\par We report the quantitative evaluation results of the pseudo labels generated with CAM in Table~\ref{tab_pseudo_label_s}. The results are evaluated on the \texttt{train} and \texttt{val} set of the VOC dataset and compared with recent WSSS methods. Since ViT is initially pretrained on ImageNet-21k, for a fair comparison with other methods, we also report the results of ToCo with the ImageNet-1k pretrained weights, \ie DeiT \cite{touvron2021training}. Table~\ref{tab_pseudo_label_s} shows that ToCo can produce higher quality pseudo labels than the competitors with both the ImageNet-1k and ImageNet-21k pretrained weights. Particularly, ToCo can significantly outperform ViT-PCM \cite{ros2022max} which also uses ViT-B$^\dagger$, even though the latter employs additional CRF post-processing.

\begin{figure*}[!t]
  \centering
  \includegraphics[width=0.8\linewidth]{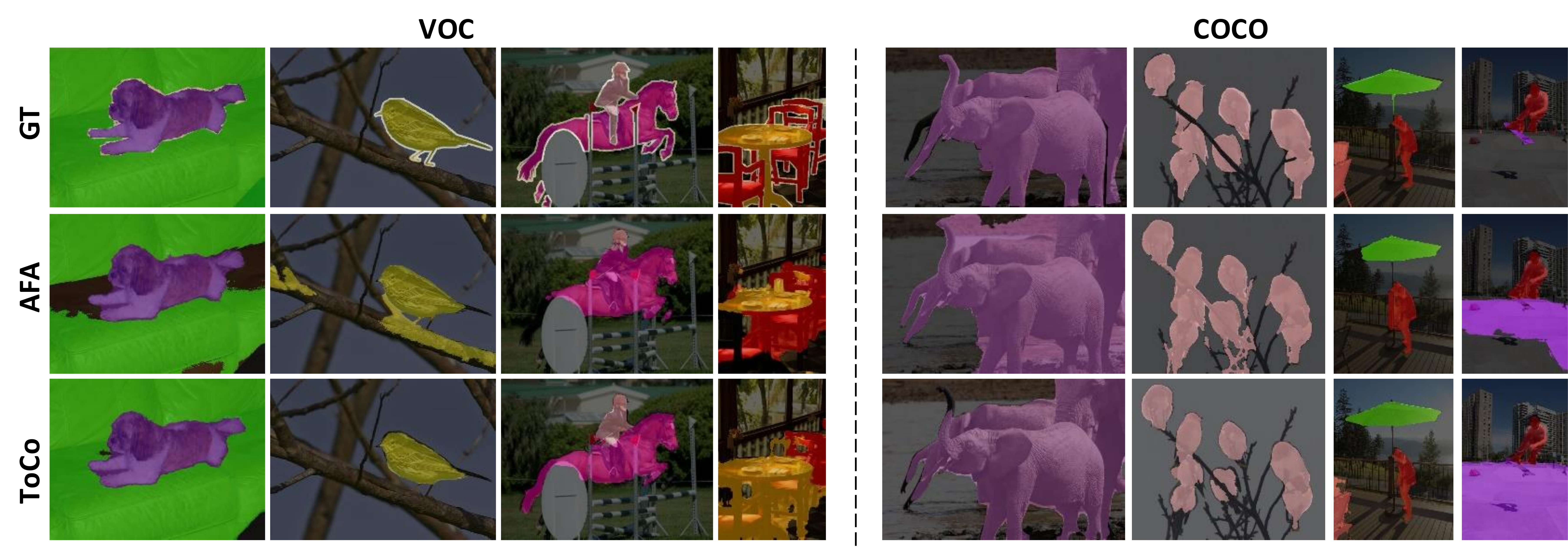}
  \caption{\textbf{Semantic segmentation results on the VOC and COCO dataset}. The last column shows a failure case.}
  \label{fig_seg}
\end{figure*}

\noindent\textbf{Semantic Segmentation Results.}
The semantic segmentation results on the VOC and COCO datasets are reported in Table~\ref{tab_sem_seg}. The proposed ToCo achieves 71.1\%, 72.2\% and 42.3\% mIoU on the VOC \texttt{val}, \texttt{test} and COCO \texttt{val} set, respectively, which largely outperform previous single-stage methods. Particularly, our single-stage ToCo achieves higher mIoU than multi-stage WSSS methods using image-level labels only. Other multi-stage methods using extra saliency maps ($\mathcal{I}+\mathcal{S}$) only slightly outperform ToCo.
\par In Figure~\ref{fig_seg}, we visualize and compare the predicted segmentation masks of ToCo, AFA \cite{ru2022learning}, and the ground truth labels. Figure~\ref{fig_seg} shows ToCo can produce more accurate segmentation masks than AFA. Our single-stage WSSS results are also very close to the ground truth.

\begin{table}[tbp]
  \centering
  \center
  \small
  \setlength{\tabcolsep}{0.008\textwidth}
  {
    \begin{tabular}{l|c|cc|c}
      \toprule
      Method                            & {Backbone}      & \texttt{val} ($\mathcal{F}$) & \texttt{val} ($\mathcal{I}$) & \textit{ratio}   \\ \midrule
      1Stgae \cite{araslanov2020single} & WR38            & 80.8                         & 62.7                         & 77.59\%          \\
      AFA \cite{ru2022learning}         & MiT-B1          & 78.7                         & 66.0                         & 83.86\%          \\
      SLRNet \cite{pan2022learning}     & WR38            & 80.8                         & 67.2                         & 83.17\%          \\ \midrule
      \rowcolor[HTML]{eaeaea}
                                        & ViT-B           & 80.5                         & 69.8                         & \textbf{86.71\%} \\
      \rowcolor[HTML]{eaeaea}
      \multirow{-2}{*}{{ToCo}}          & ViT-B$^\dagger$ & \textbf{82.3}                & \textbf{71.1}                & {86.39\%}        \\ \bottomrule
    \end{tabular}
  }
  \caption{\textbf{The fully-supervised counterparts of single-stage WSSS methods.} $\mathcal{F}$ / $\mathcal{I}$ denotes the pixel- / image-level supervision. $\dagger$ denotes using ImageNet-21k \cite{ridnik2021imagenet} pretrained parameters.}
  \label{tab_full_sup}%

  \vspace{-6mm}
\end{table}

\subsection{Ablation and Analysis}
\label{sec_ablation}

\noindent\textbf{Ablation.} To investigate the impact of the proposed PTC and CTC, in Table~\ref{tab_ablation}, we report the performance of the generated final CAM ($\textbf{M}$), auxiliary CAM ($\textbf{M}^{m}$) and the semantic segmentation results ($Seg.$). The results are evaluated on the VOC \texttt{val} set.
\par We first show that due to over-smoothing, training a baseline ViT with the classification loss (\ie $\mathcal{L}_{cls}$) cannot produce reasonable CAM. Besides, adding an auxiliary classifier (\ie $\mathcal{L}_{cls} + \mathcal{L}^{m}_{cls}$) can help derive the auxiliary CAM from the intermediate layer, but it still cannot address the over-smoothing issue. The proposed PTC module can finely address this issue and significantly improve the quality of the generated CAM $\textbf{M}$, \ie from 27.9\% mIoU to 62.5\% mIoU. Table~\ref{tab_ablation} also shows that the improvements in the final layer also benefit the intermediate CAM $\textbf{M}^{m}$, improving $\textbf{M}^{m}$ to 57.8\% mIoU. The CTC module further improves the quality of pseudo labels by 4.7\% mIoU. It's noted that the segmentation loss also improves the CAM's quality, since we use PAR \cite{ru2022weakly} to refine the pseudo labels which enforce the pseudo labels to align better with the low-level object boundaries. The regularization term $\mathcal{L}_{reg}$ also brings improvements to the semantic segmentation results.

\begin{table}[tbp]
  \centering

  \center
  \small
  \setlength{\tabcolsep}{0.008\textwidth}
  {
    \begin{tabular}{l|cc|cc|cc|c}
      \toprule
      {Method}                                                       & PTC        & CTC        & $\mathcal{L}_{seg}$ & $\mathcal{L}_{reg}$ & $\textbf{M}$  & $\textbf{M}^{m}$ & $Seg.$        \\\midrule
      $\mathcal{L}_{cls}$                                            &            &            &                     &                     & 27.8          & --               & --            \\
      $\mathcal{L}_{cls} + \mathcal{L}^{m}_{cls}$                    &            &            &                     &                     & 27.9          & 53.8             & --            \\ \midrule
      \multirow{4}{*}{{$\mathcal{L}_{cls} + \mathcal{L}^{m}_{cls}$}} & \checkmark &            &                     &                     & 62.5          & 57.8             & --            \\
                                                                     & \checkmark & \checkmark &                     &                     & 67.2          & 60.7             & --            \\
                                                                     & \checkmark & \checkmark & \checkmark          &                     & 69.9          & 61.2             & 66.6          \\
                                                                     & \checkmark & \checkmark & \checkmark          & \checkmark          & \textbf{70.5} & \textbf{62.5}    & \textbf{68.1} \\ \bottomrule
    \end{tabular}
  }
  \caption{\textbf{Ablation Study.} $\textbf{M}^{m}$: auxiliary CAM from the intermediate layer; $\textbf{M}$: CAM from the final layer; $Seg.$: semantic segmentation results.}
  \label{tab_ablation}%

  \vspace{-6mm}
\end{table}

\begin{table*}[!tbp]

  \centering
  \small
  \begin{subtable}{0.33\textwidth}
    \setlength{\tabcolsep}{3mm}
    \centering
    \begin{tabular}{l|ccc}
      \toprule
      Block & $\mathbf{M}$ & $\mathbf{M}^m$ & $Seg.$ \\ \midrule
      \#8   & 63.5         & 48.1           & 60.5   \\
      \#9   & 67.6         & 55.1           & 64.8   \\
      \rowcolor[HTML]{eaeaea}
      \#10  & 70.5         & 62.5           & 68.1   \\
      \#11  & 43.1         & 45.2           & 40.3   \\ \bottomrule
    \end{tabular}
    \caption{\textbf{Index of auxiliary block}.}
    \label{tab_block_index}
  \end{subtable}
  \begin{subtable}{0.33\textwidth}
    \setlength{\tabcolsep}{3mm}
    \centering
    \begin{tabular}{l|cc}
      \toprule
      Size    & $\mathbf{M}$ & $Seg.$ \\ \midrule
      64$^2$  & 68.1         & 65.5   \\
      80$^2$  & 69.3         & 67.3   \\
      \rowcolor[HTML]{eaeaea}
      96$^2$  & 70.5         & 68.1   \\
      112$^2$ & 69.4         & 67.0   \\ \bottomrule
    \end{tabular}
    \caption{\textbf{Crop size of local view}.}
    \label{tab_local_size}
  \end{subtable}
  \begin{subtable}{0.33\textwidth}
    \setlength{\tabcolsep}{4mm}
    \centering
    \begin{tabular}{l|cc}
      \toprule
      Momentum & $\mathbf{M}$ & $Seg.$ \\ \midrule
      0        & 69.2         & 66.9   \\
      0.1      & 69.0         & 67.1   \\
      \rowcolor[HTML]{eaeaea}
      0.9      & 70.5         & 68.1   \\
      0.99     & 68.7         & 66.6   \\ \bottomrule
    \end{tabular}
    \caption{\textbf{Momentum in EMA}.}
    \label{tab_ctc_momentum}
  \end{subtable}

  \caption{\textbf{Impact of hyper-parameters.} The performance is evaluated on the VOC \texttt{val} set. The default settings are marked in \colorbox{baselinecolor}{gray}.}
  \label{tab_params}
  \vspace{-4mm}
\end{table*}

\noindent\textbf{Analysis of PTC.} To demonstrate that PTC addresses the over-smoothing issue well, in Figure~\ref{fig_voc_cos_sim}, we compare the average pairwise cosine similarity of patch tokens of ViT with and without PTC. We show that in the late layers, the average cosine similarity with PTC is remarkably lower than the baseline. Note that the cosine similarity values decrease after the $10^{th}$ layer, since we choose the $10^{th}$ layer to produce the auxiliary labels and supervise the final CAM. We also visualize the generated CAM with and without PTC in Figure~\ref{fig_toco_cam}. Figure~\ref{fig_toco_cam} shows PTC helps produce reasonable CAM, which coincides with Table~\ref{tab_ablation}.
\begin{figure}[tbp]
  \centering
  \includegraphics[width=0.9\linewidth]{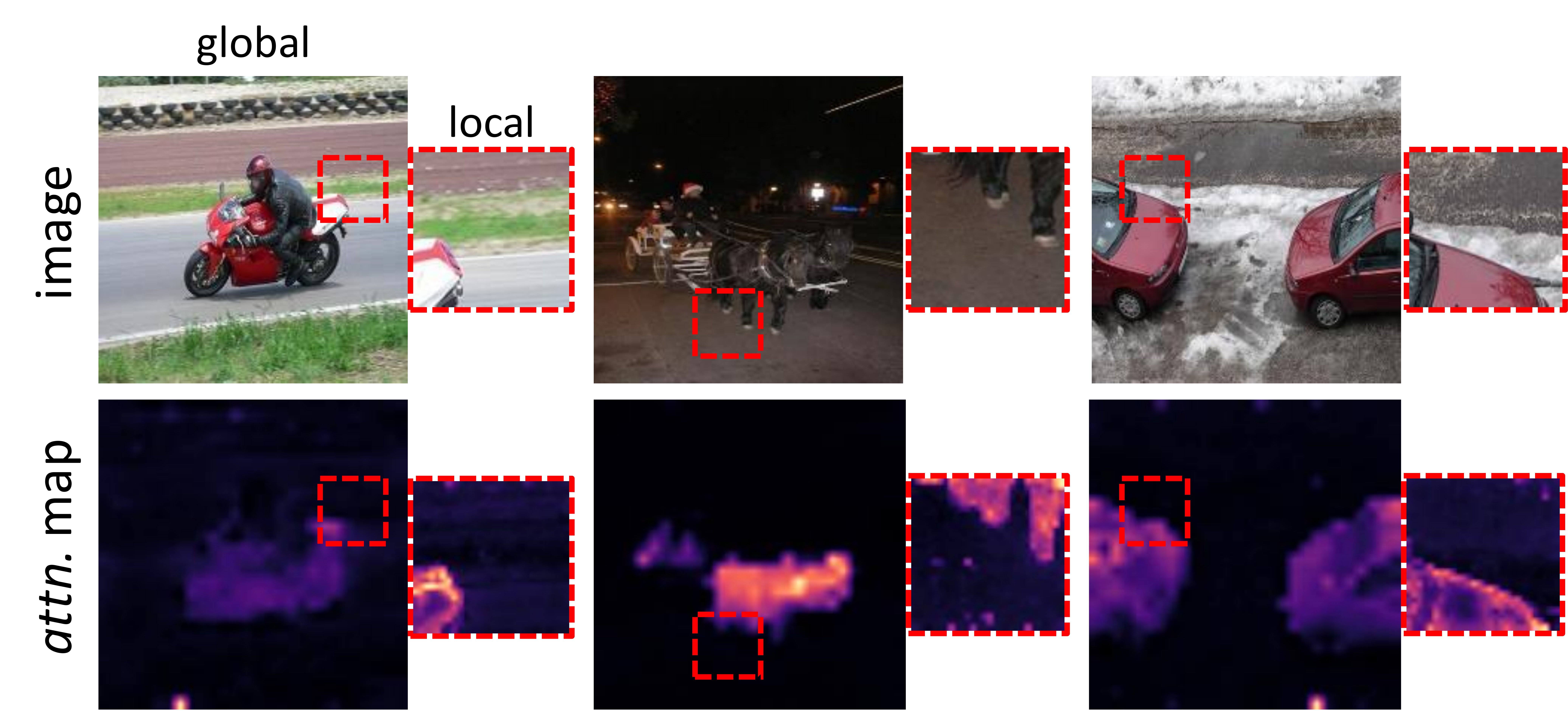}
  \caption{\textbf{Visualization of the attention map of class token \wrt patch tokens}. The brighter region indicates a larger attention value. \textit{Left}: the global view image in CTC; \textit{Right}: the local view image randomly cropped from the global view in CTC.}
  \label{fig_cls_token_maps}
  \vspace{-6mm}
\end{figure}

\noindent\textbf{Analysis of CTC.} The motivation of CTC is to encourage the local-to-global consistency of semantic objects. In Figure~\ref{fig_cls_token_maps}, we visualize the attention map of class token \wrt other patch tokens in ToCo. We show that for both the global and local view, the class token finely captures the foreground object information. Moreover, the class token of the local view can learn the less salient regions that are usually ignored in the global view, which fulfils our intention. In Figure~\ref{fig_toco_cam}, we also visualize the generated CAM with CTC (ViT+PTC+CTC) and without CTC (ViT+PTC). Figure~\ref{fig_toco_cam} also shows the proposed CTC can help to activate the less discriminative regions, which accounts for the quantitative improvements in Table~\ref{tab_ablation}.

A possible concern is that the objective of CTC may be limited when the global image includes multiple semantics but the local images only include partial semantics. However, as shown in Figure~\ref{fig_cls_token_maps}, the class token can capture multi-class semantic information. Therefore, when a local image only covers one semantic class, CTC can still enforce local-global representation consistency, though it's not a perfect objective. Moreover, considering the whole dataset, a semantic class usually co-occurs with multiple other classes, so the optimization to different classes can also be neutralized. Figure~\ref{fig_toco_cam} shows CTC works well on images with multiple semantics.

\noindent\textbf{Fully-Supervised Counterparts.} The single-stage methods in Table~\ref{tab_sem_seg} use different backbones. To ensure the fairness of comparison, we report their upper bound performance on the VOC \texttt{val} set, \ie the performance of their fully-supervised counterparts. Table~\ref{tab_full_sup} shows that using only image-level labels as supervision, ToCo with ViT-B as backbone achieves 69.8\% mIoU, which is 86.71\% of its upper bound performance. In further, ToCo with ViT-B$^\dagger$ as backbone can achieve higher performance. Compared to ToCo, the previous methods, AFA \cite{ru2022learning} and SLRNet \cite{pan2022learning} only achieve 83.86\% and 83.17\% of their fully-supervised counterparts, respectively. Particularly, WideResNet38 (WR38) can achieve comparable performance with ViT-B under full pixel-level supervision. However, in the WSSS experiments, ToCo can remarkably outperform SLRNet.

\noindent\textbf{Auxiliary Classifier.} Table~\ref{tab_block_index} shows the impact of using different blocks to produce the auxiliary CAM. Usually, the shallow blocks cannot capture high-level semantics while the late blocks encounter the over-smoothing issue. For the ViT-B backbone used in our experiments ($12$ Transformer blocks), we empirically observe that appending the auxiliary classifier in the $10^{th}$ block is a preferred choice.

\noindent\textbf{Local Crop Size.} The CTC module compares the representations of local and global view images to encourage consistency between the discriminative and less-discriminative object regions. Intuitively, a smaller local image may fail to include the foreground objects, mismatching the objective. On the contrary, a larger local image could contain too many discriminative object regions, also affecting distinguishing the uncertain regions. In Table~\ref{tab_local_size}, we show that cropping local images with a size of $96^2$ can achieve the best performance.

\noindent\textbf{EMA in CTC.} We use EMA to update the parameters of the global projection head $\mathcal{P}^g$ in Figure~\ref{fig_toco}. In Table~\ref{tab_ctc_momentum}, we report the impact of the momentum value $\rho$ in the EMA process. Table~\ref{tab_ctc_momentum} shows $\rho=0.9$ is the best choice but other values can also yield favorable performance. 

\section{Conclusion}
In this work, we aim to address the over-smoothing issue of ViT and further exploit its virtue for WSSS. Specifically, we first design a Patch Token Contrast module (PTC). PTC contrasts the final patch representations with knowledge extracted from the intermediate layer, which is empirically proved to counter the over-smoothing issue well. Inspired by the observation that the class token in ViT can capture the high-level semantics, we further propose a Class Token Contrast module (CTC) to contrast the class tokens of the local and global images, which can facilitate the representation consistency of the integral object regions. We plug ToCo into the single-stage WSSS framework and conduct extensive experiments on the VOC and COCO datasets. The experimental results show that ToCo can significantly outperform other competitors.

  {\small
    \bibliographystyle{ieee_fullname}
    \bibliography{main.bbl}
  }

\clearpage

\section{Additional Results}

\subsection{Backbone with ViT Variants}
\par
\begin{figure*}[tbp]
  \centering

  \begin{minipage}{0.35\textwidth}
    \centering

    \begin{tabular}{l|l|cc}
      \toprule
      \multicolumn{2}{c|}{\textbf{ViT-S}} & CAM               & $Seg.$                        \\ \midrule
      \multirow{2}{*}{\texttt{train}}     & \textit{w/o} ToCo & 27.3          & --            \\
                                          & \textit{w/} ToCo  & \textbf{69.1} & \textbf{68.1} \\ \midrule
      \multirow{2}{*}{\texttt{val}}       & \textit{w/o} ToCo & 27.6          & --            \\
                                          & \textit{w/} ToCo  & \textbf{68.2} & \textbf{65.2} \\ \bottomrule
    \end{tabular}
  \end{minipage}
  \begin{minipage}{0.55\textwidth}
    \centering
    \includegraphics[width=0.95\textwidth]{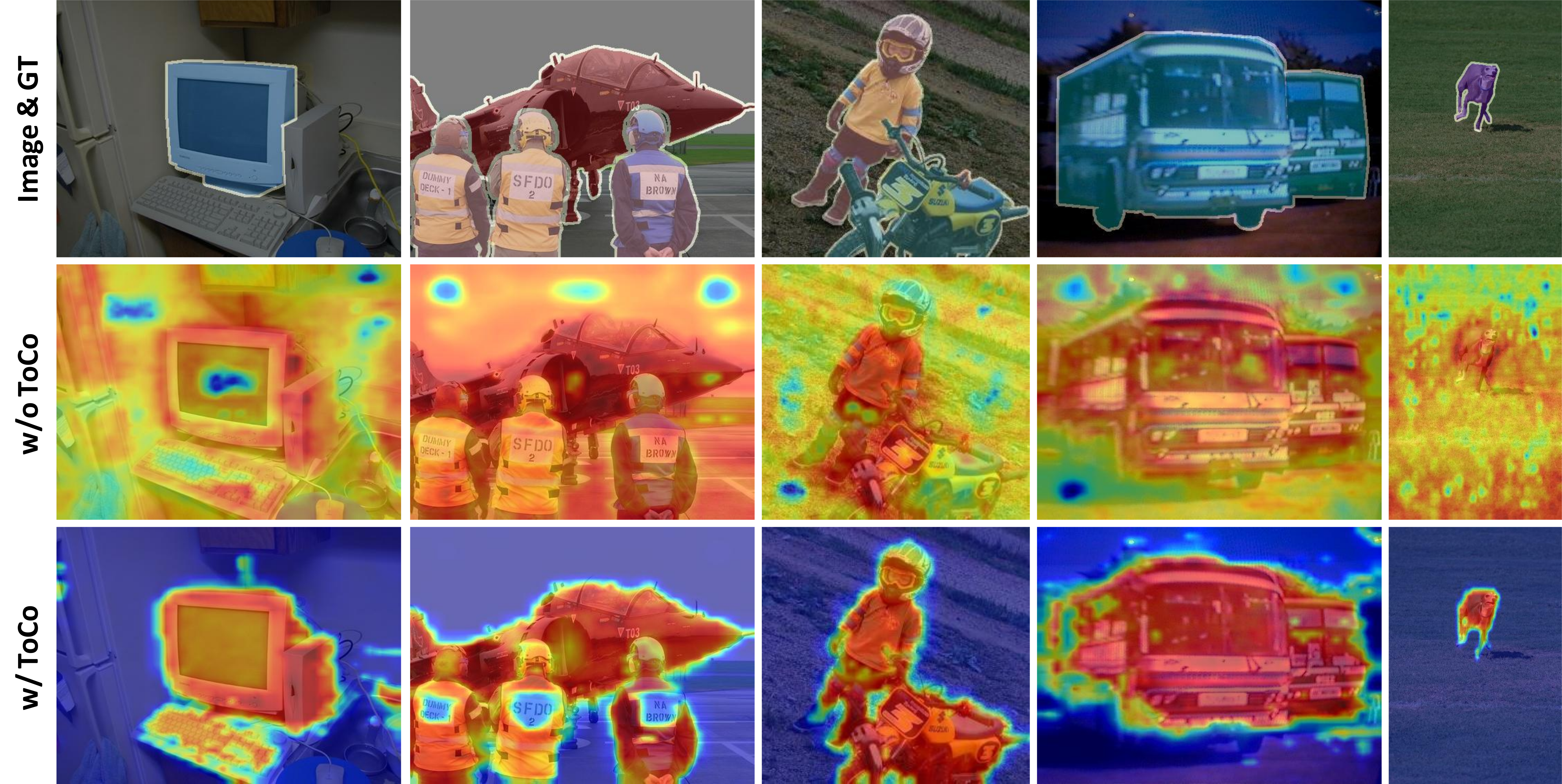}
  \end{minipage}
  \caption{\textbf{Evaluation of the generated CAM and semantic segmentation results with ViT-S}. The results are evaluated on the VOC dataset.}
  \label{fig_vit_s}
  \vspace{-0.4cm}
\end{figure*}

\begin{figure*}[tbp]
  \centering

  \begin{minipage}{0.35\textwidth}
    \centering

    \begin{tabular}{l|l|cc}
      \toprule
      \multicolumn{2}{c|}{\textbf{ViT-L}$^\dagger$} & CAM               & $Seg.$                        \\ \midrule
      \multirow{2}{*}{\texttt{train}}               & \textit{w/o} ToCo & 25.3          & --            \\
                                                    & \textit{w/} ToCo  & \textbf{73.8} & \textbf{74.2} \\ \midrule
      \multirow{2}{*}{\texttt{val}}                 & \textit{w/o} ToCo & 25.6          & --            \\
                                                    & \textit{w/} ToCo  & \textbf{72.6} & \textbf{71.2} \\ \bottomrule
    \end{tabular}
  \end{minipage}
  \begin{minipage}{0.55\textwidth}
    \centering
    \includegraphics[width=0.95\textwidth]{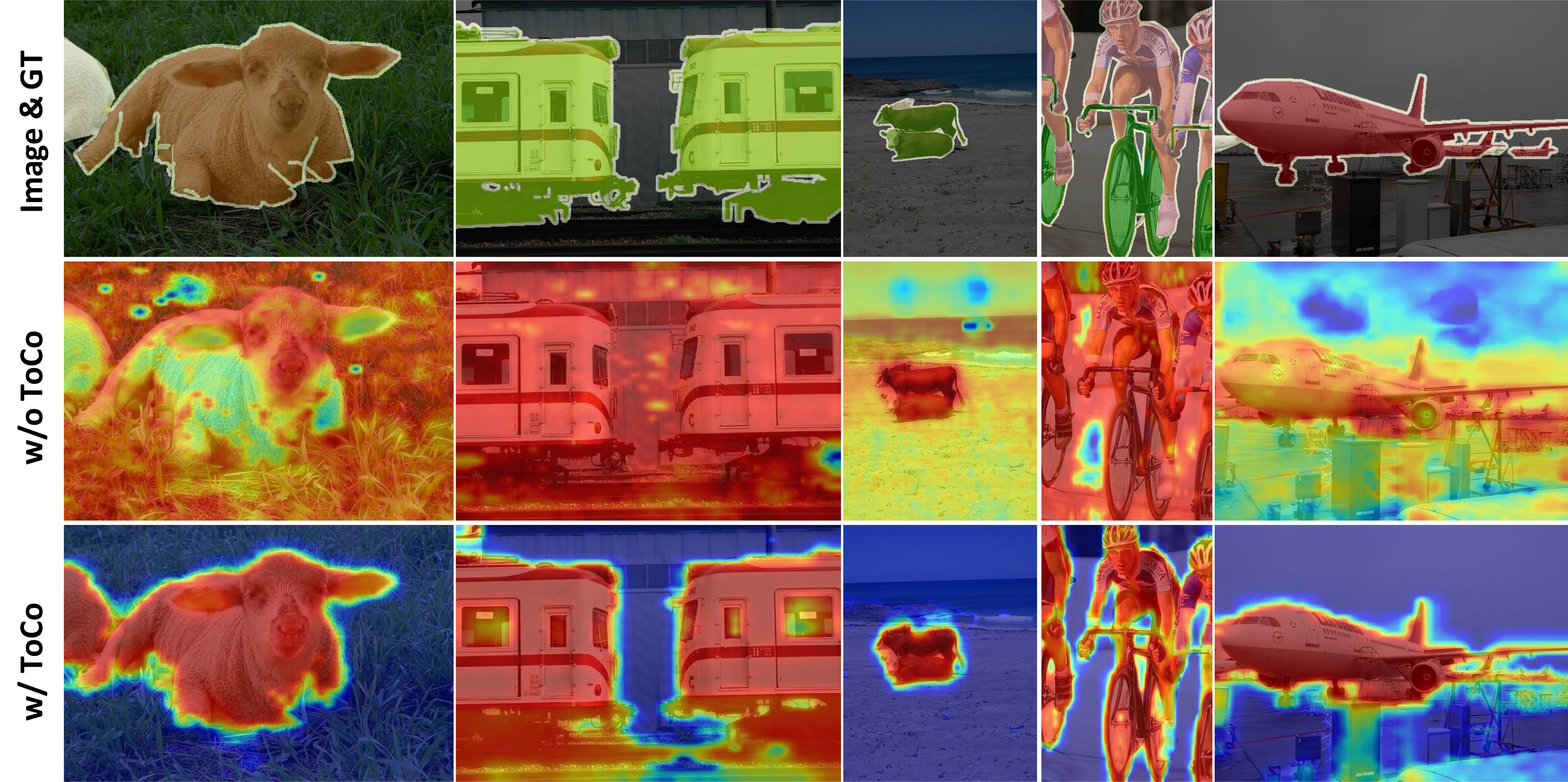}
  \end{minipage}

  \caption{\textbf{Evaluation of the generated CAM and semantic segmentation results with ViT-L$^\dagger$}. The results are evaluated on the VOC dataset.}
  \label{fig_vit_l}
  \vspace{-0.4cm}
\end{figure*}

 In the previous experiments, we mainly conducted experiments with ViT-B as the backbone. In Figure~\ref{fig_vit_s} and Figure~\ref{fig_vit_l}, we report the evaluation of the generated CAM and semantic results with ViT using other configurations (ViT-S, ViT-L$^\dagger$ \cite{dosovitskiy2020image}). ViT-S and ViT-L consist of 12 and 24 Transformer blocks, respectively. We show that other backbones also encounter the over-smoothing issue and the proposed ToCo can finely address it. Specifically, without the proposed ToCo, the generated CAM typically activates all image regions, and the semantic segmentation results also perform badly. In a contrast, the proposed ToCo finely addresses the over-smoothing issue and promotes the semantic segmentation performance to 65.2\% and 71.2\% mIoU with ViT-S and ViT-L as the backbone, respectively. 

\subsection{Hyper-parameters}
We report the impact of other hyper-parameters in this section.

\noindent\textbf{Background Thresholds.} In Table~\ref{tab_bkg_thres}, we report the impact of background thresholds to differentiate the foreground, background, and uncertain regions. We show the combination of $\beta_h=0.7$ and $\beta_l=0.25$ can achieve the best performance.

\noindent\textbf{Temperature Factors.} Table~\ref{tab_ctc_temp} presents the performance \wrt the temperature factor $\tau$ in Equation~(4). $\tau$ control the sharpness of the logits. In Table~\ref{tab_ctc_temp}, we observe that $\tau_g$=0.5 yields the best performance, while other values can also achieve favorable performance.

\noindent\textbf{Loss Weights.} Table~\ref{tab_loss_weights} reports the analysis of the weights of loss terms. The combination of ($\lambda_1$ = 0.2, $\lambda_2$ = 0.5, $\lambda_3$ = 0.1) can produce the best semantic segmentation results.

\subsection{Setting of $\mathcal{L}_{ptc}$}
\par In the PTC module, due to the observation that the cosine similarities of patch tokens are usually positive values, as indicated in \cite{gong2021vision,chen2022principle}, we use the absolute cosine similarity instead of the origin cosine similarity in $\mathcal{L}_{ptc}$. In Table~\ref{tab_ptc_loss}, we report the evaluation of pseudo labels $\mathbf{M}$ and semantic segmentation results $Seg.$
\begin{table}[hbp]
  \centering
  \center
  \small
  \setlength{\tabcolsep}{0.04\textwidth}
  \begin{tabular}{l|cc}
    \toprule
                            & $\mathbf{M}$ & $Seg.$ \\ \midrule
    $\mathtt{CosSim}$       & 36.5         & 29.6   \\
    $\mathtt{ReLu(CosSim)}$ & 63.0         & 60.1   \\
    $\mathtt{Abs(CosSim)}$  & 70.5         & 68.1   \\ \bottomrule
  \end{tabular}
  \caption{\textbf{The impact of CosSim in $\mathcal{L}_{ptc}$}.}
  \label{tab_ptc_loss}
  \vspace{-4mm}
\end{table}

\par Table~\ref{tab_ptc_loss} shows that directly minimizing the cosine similarity ($\mathtt{CosSim}$) cannot produce satisfactory results. A possible reason is that two patch tokens with negative similarities are still correlated. When ignoring the negative parts ($\mathtt{ReLu(CosSim)}$), the results are remarkably promoted. Finally, using the absolute cosine similarity ($\mathtt{Abs(CosSim)}$) can finely optimize the positive and negative parts and yield the best results.

\begin{table*}[!tbp]

  \centering
  \small

  \begin{subtable}{0.33\textwidth}
    \setlength{\tabcolsep}{6pt}
    \centering
    \begin{tabular}{cc|cccc}
      \toprule
      \multicolumn{2}{l|}{\multirow{2}{*}{}}          & \multicolumn{4}{c}{$\beta_l$}                                                                \\ \cmidrule{3-6}
      \multicolumn{2}{l|}{}                           & 0.15                          & 0.2  & \colorbox{baselinecolor}{0.25} & 0.3                  \\ \midrule
      \multicolumn{1}{l|}{\multirow{5}{*}{$\beta_h$}} & 0.6                           & --   & 55.1                           & 62.9          & 66.7 \\
      \multicolumn{1}{l|}{}                           & 0.65                          & 51.2 & 61.5                           & 67.2          & 66.3 \\
      \multicolumn{1}{l|}{}                           & \colorbox{baselinecolor}{0.7} & 56.2 & 65.5                           & \textbf{68.1} & 67.1 \\
      \multicolumn{1}{l|}{}                           & 0.75                          & 63.0 & 65.6                           & 66.3          & 64.3 \\ \bottomrule
    \end{tabular}
    \caption{\textbf{Background thresholds}. }
    \label{tab_bkg_thres}%
  \end{subtable}
  \begin{subtable}{0.33\textwidth}
    \setlength{\tabcolsep}{6pt}
    \centering
    \begin{tabular}{l|cccc}
      \toprule
      $\tau$ & 0.1  & 0.2  & \colorbox{baselinecolor}{0.5} & 0.8  \\ \midrule
      $Seg.$ & 66.0 & 67.2 & \textbf{68.1}                 & 67.3 \\ \bottomrule
    \end{tabular}
    \caption{\textbf{Temperatures}.}
    \label{tab_ctc_temp}
  \end{subtable}
  \begin{subtable}{0.33\textwidth}
    \setlength{\tabcolsep}{6pt}
    \centering
    \begin{tabular}{l|cccc}
      \toprule
      $\lambda_1$ & 0.05 & 0.1  & \colorbox{baselinecolor}{0.2} & 0.5  \\ \midrule
      $Seg.$      & 64.2 & 67.0 & \textbf{68.1}                 & 65.4 \\\bottomrule
    \end{tabular}\vspace{1mm}
    \begin{tabular}{l|cccc}
      \toprule
      $\lambda_2$ & 0.1  & 0.2  & \colorbox{baselinecolor}{0.5} & 0.8  \\ \midrule
      $Seg.$      & 65.1 & 65.9 & \textbf{68.1}                 & 67.1 \\\bottomrule
    \end{tabular}\\\vspace{1mm}
    \begin{tabular}{l|cccc}
      \toprule
      $\lambda_3$ & 0.05 & \colorbox{baselinecolor}{0.1} & 0.2  & 0.5  \\ \midrule
      $Seg.$      & 67.8 & \textbf{68.1}                 & 67.6 & 66.0 \\ \bottomrule
    \end{tabular}
    \caption{\textbf{Loss Weights}.}
    \label{tab_loss_weights}
  \end{subtable}

  \caption{\textbf{Impact of hyper-parameters.} The performance is evaluated on the VOC \texttt{val} set. The default settings are marked in \colorbox{baselinecolor}{gray}.}
  \label{supp_tab_params}
  \vspace{-2mm}
\end{table*}

\subsection{Additional Quantitative/Qualitative Results}
\noindent\textbf{Per-Class Results.} We report the per-class semantic segmentation results on the VOC $val$ set in Table~\ref{tab_miou}. Table~\ref{tab_miou} shows that the proposed ToCo can achieve the highest accuracy in most semantic classes.
\begin{table*}[!tp]
  \centering
  \footnotesize
  \setlength{\tabcolsep}{3pt}%
  \begin{tabular}{l|ccccccccccccccccccccc|c}
    \toprule
                                          & \textbf{\rotatebox[origin=c]{70}{bkg}} & \textbf{\rotatebox[origin=c]{70}{aero}} & \textbf{\rotatebox[origin=c]{70}{bicycle}} & \textbf{\rotatebox[origin=c]{70}{bird}} & \textbf{\rotatebox[origin=c]{70}{boat}} & \textbf{\rotatebox[origin=c]{70}{bottle}} & \textbf{\rotatebox[origin=c]{70}{bus}} & \textbf{\rotatebox[origin=c]{70}{car}} & \textbf{\rotatebox[origin=c]{70}{cat}} & \textbf{\rotatebox[origin=c]{70}{chair}} & \textbf{\rotatebox[origin=c]{70}{cow}} & \textbf{\rotatebox[origin=c]{70}{table}} & \textbf{\rotatebox[origin=c]{70}{dog}} & \textbf{\rotatebox[origin=c]{70}{horse}} & \textbf{\rotatebox[origin=c]{70}{motor}} & \textbf{\rotatebox[origin=c]{70}{person}} & \textbf{\rotatebox[origin=c]{70}{plant}} & \textbf{\rotatebox[origin=c]{70}{sheep}} & \textbf{\rotatebox[origin=c]{70}{sofa}} & \textbf{\rotatebox[origin=c]{70}{train}} & \textbf{\rotatebox[origin=c]{70}{tv}} & \textbf{\rotatebox[origin=c]{70}{mIoU}} \\ \midrule
    {{1Stage} \cite{araslanov2020single}} & 88.7                                   & 70.4                                    & {35.1}                                     & 75.7                                    & 51.9                                    & 65.8                                      & 71.9                                   & 64.2                                   & 81.1                                   & 30.8                                     & 73.3                                   & 28.1                                     & 81.6                                   & 69.1                                     & 62.6                                     & 74.8                                      & 48.6                                     & 71.0                                     & 40.1                                    & \textbf{68.5}                            & {64.3}                                & 62.7                                    \\
    AFA \cite{ru2022learning}             & {89.9}                                 & {79.5}                                  & 31.2                                       & \textbf{80.7}                           & {67.2}                                  & 61.9                                      & {81.4}                                 & 65.4                                   & {82.3}                                 & 28.7                                     & {83.4}                                 & 41.6                                     & {82.2 }                                & {75.9}                                   & 70.2                                     & 69.4                                      & {53.0}                                   & {85.9}                                   & \textbf{44.1}                           & 64.2                                     & 50.9                                  & {66.0}                                  \\ \midrule
    {ToCo}                                & 89.9                                   & \textbf{81.8}                           & 35.4                                       & 68.1                                    & \textbf{62.0}                           & 76.6                                      & 83.6                                   & 80.4                                   & 87.7                                   & 24.5                                     & \textbf{88.1}                          & 54.9                                     & \textbf{87.0}                          & \textbf{84.0}                            & 76.0                                     & 68.2                                      & \textbf{65.6}                            & 85.8                                     & 42.4                                    & 57.7                                     & \textbf{65.6}                         & 69.8                                    \\
    ToCo$^\dagger$                        & \textbf{91.1}                          & 80.6                                    & \textbf{48.7}                              & 68.6                                    & 45.4                                    & \textbf{79.6}                             & \textbf{87.4}                          & \textbf{83.3}                          & \textbf{89.9}                          & \textbf{35.8}                            & 84.7                                   & \textbf{60.5}                            & 83.7                                   & 83.2                                     & \textbf{76.8}                            & \textbf{83.0}                             & 56.6                                     & \textbf{87.9}                            & 43.5                                    & 60.5                                     & 63.1                                  & \textbf{71.1}                           \\ \bottomrule
  \end{tabular}
  \vspace{-2mm}
  \caption{Evaluation and comparison of the semantic segmentation results in mIoU on the $val$ set. $\dagger$ denotes using ImageNet-21k \cite{ridnik2021imagenet} pretrained weights.}
  \label{tab_miou}
\end{table*}

\begin{figure*}[t]
  \centering
  \includegraphics[width=0.85\linewidth]{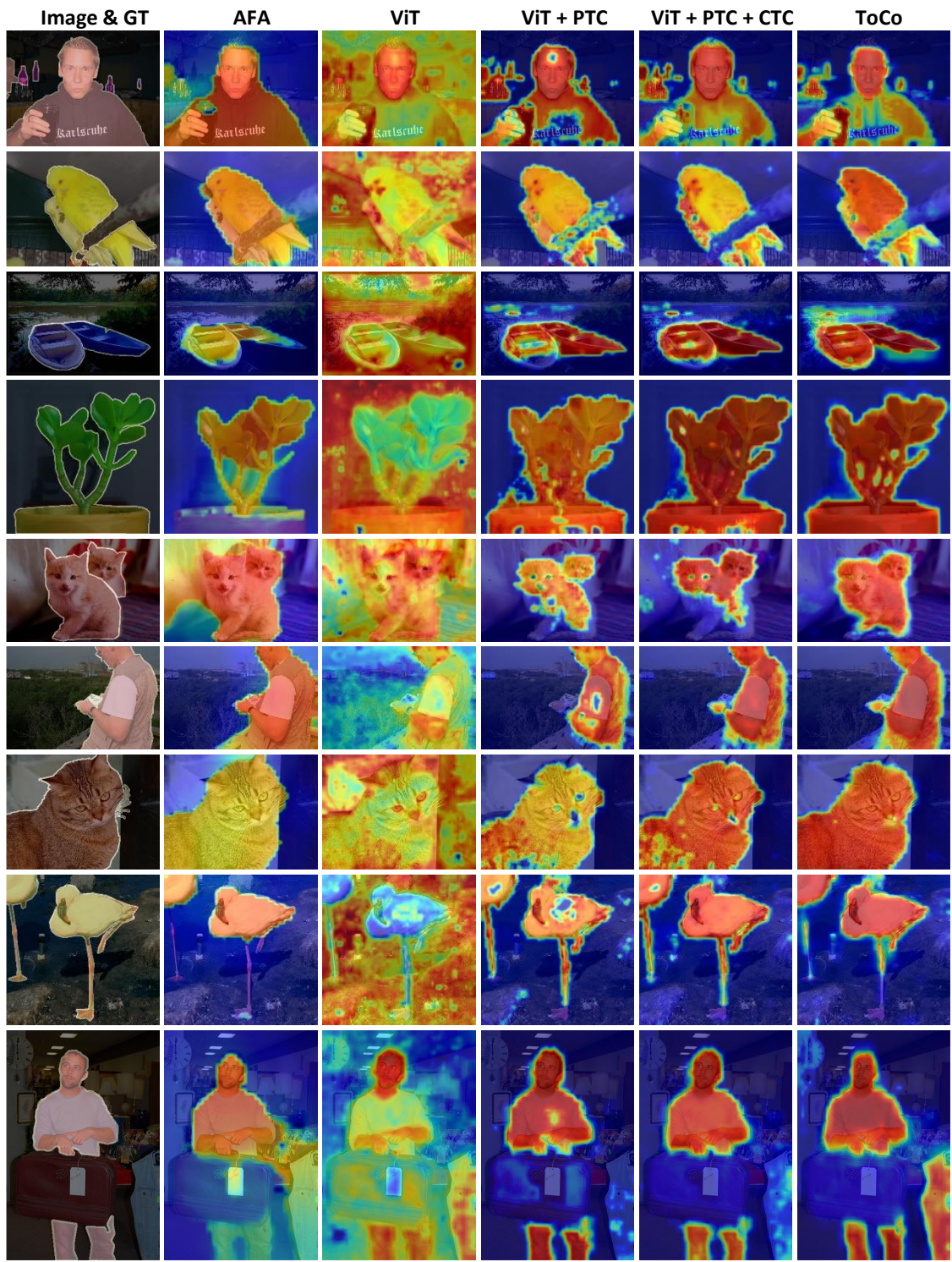}
  \caption{\textbf{Visualization of CAM.} From left to right, the CAM is generated with AFA \cite{ru2022learning}, ViT baseline, ViT with PTC, ViT with PTC and CTC, and the proposed ToCo. }
  \label{supp_fig_toco_cam}
  \vspace{-4mm}
\end{figure*}

\noindent\textbf{Pseudo Labels.} We present the generated CAM in Figure~\ref{supp_fig_toco_cam}. Figure~\ref{supp_fig_toco_cam} demonstrates that the proposed PTC and CTC can address the over-smoothing issue and further distinguish the uncertain regions, respectively. Besides, ToCo can generate better pseudo labels than the recent state-of-the-art single-stage method, AFA \cite{ru2022learning}.

\begin{figure*}[!t]
  \centering
  \includegraphics[width=0.9\linewidth]{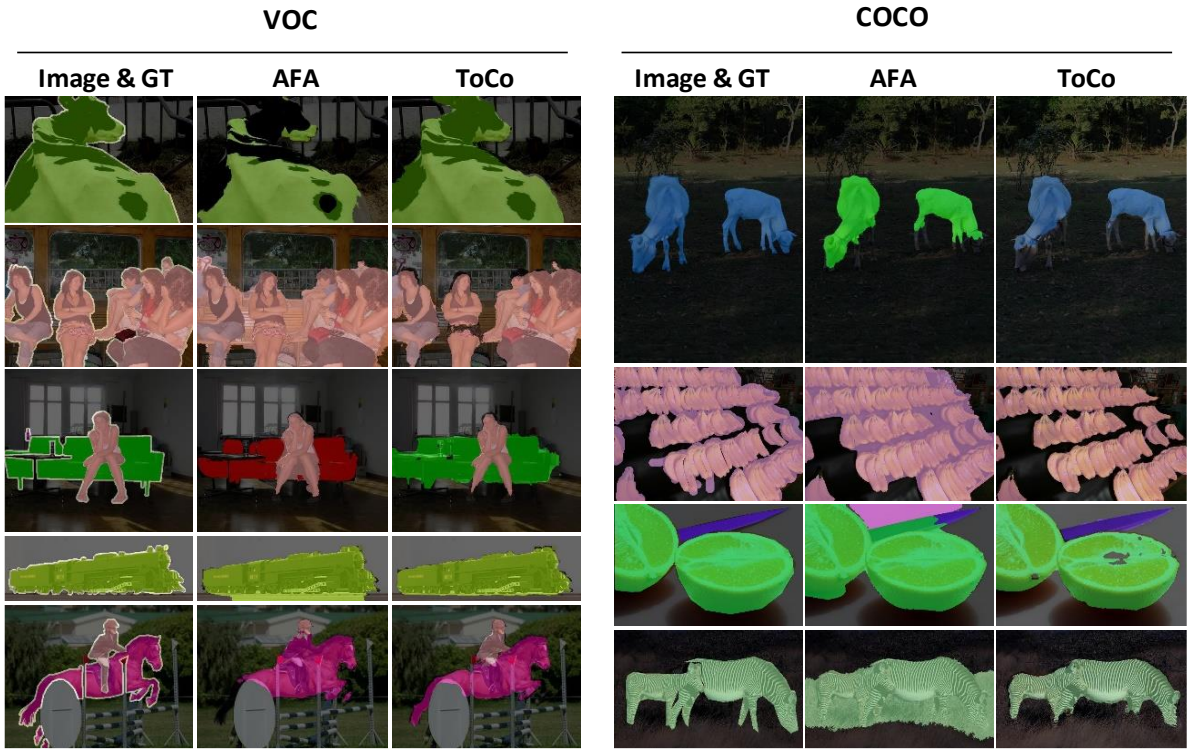}
  \caption{\textbf{Semantic segmentation results on the VOC and COCO dataset}.}
  \label{supp_fig_seg}
\end{figure*}

\noindent\textbf{Semantic Segmentation Results.} The qualitative semantic segmentation in Figure~\ref{supp_fig_seg} shows that ToCo can surpass AFA \cite{ru2022learning} and achieve close results with the ground-truth.

\begin{figure*}[tbp]
  \centering
  \includegraphics[width=0.95\linewidth]{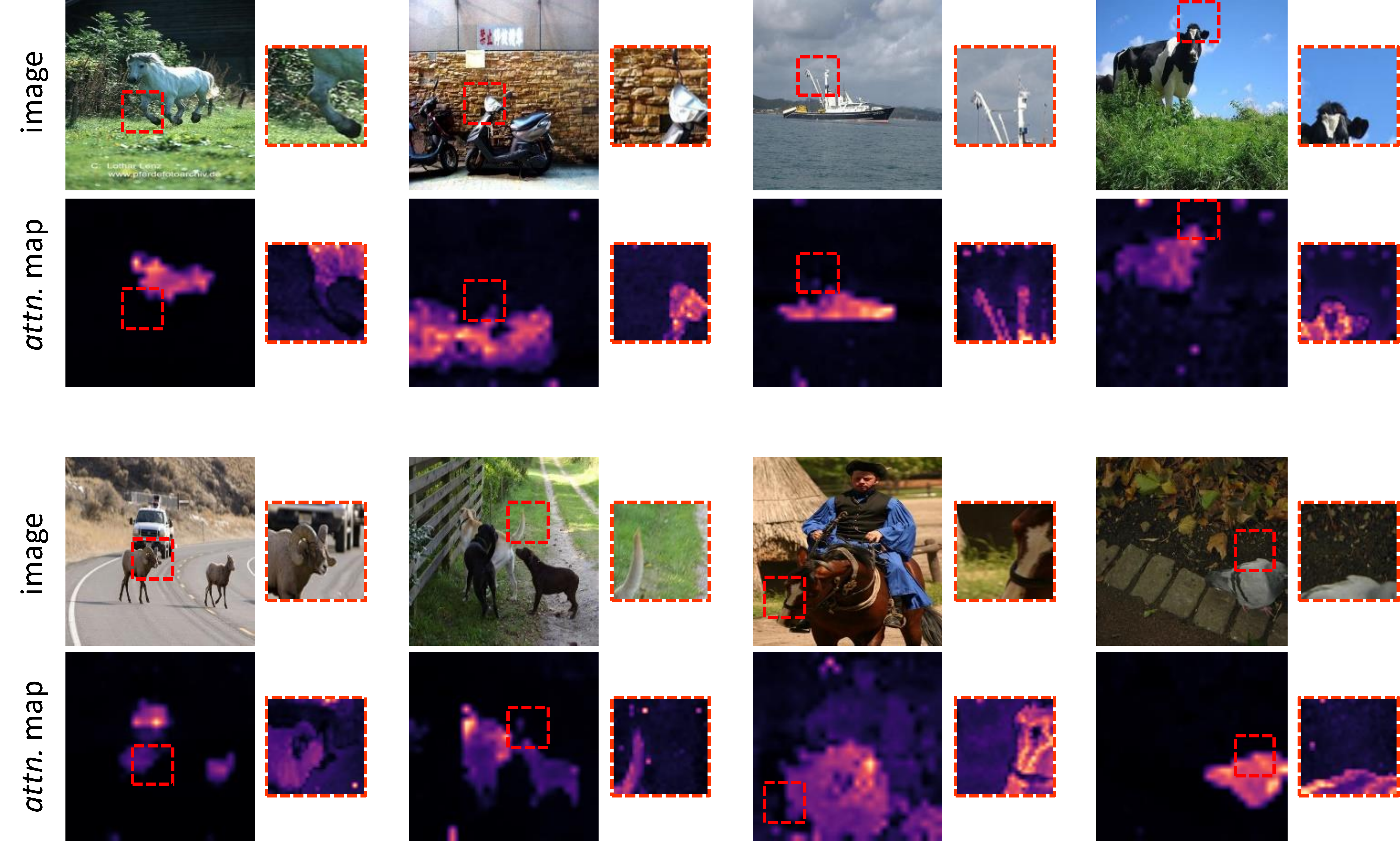}
  \caption{\textbf{Visualization of the attention map of class token \wrt patch tokens}. The brighter region indicates a larger attention value. \textit{Left}: the global view image in CTC; \textit{Right}: the local view image randomly cropped from the global view in CTC.}
  \label{supp_fig_cls_token_maps}
  \vspace{-4mm}
\end{figure*}

\noindent\textbf{Attention Maps of Class Token.} In Figure~\ref{supp_fig_cls_token_maps}, we visualize more attention maps of class token \wrt other patch tokens. Figure~\ref{supp_fig_cls_token_maps} shows the global view can discover most object regions but ignore some uncertain local regions, which can be activated in the local view. By contrasting the class tokens of global and local views in the CTC module, the representation of the integral regions can be more consistent.

\end{document}